\documentclass{article}

\usepackage{arxiv}

\usepackage[utf8]{inputenc} % allow utf-8 input
\usepackage[T1]{fontenc}    % use 8-bit T1 fonts
\usepackage{hyperref}       % hyperlinks
\usepackage{url}            % simple URL typesetting
\usepackage{booktabs}       % professional-quality tables
\usepackage{amsfonts}       % blackboard math symbols
\usepackage{nicefrac}       % compact symbols for 1/2, etc.
\usepackage{microtype}      % microtypography
\usepackage{lipsum}		% Can be removed after putting your text content
\usepackage{graphicx}
\usepackage{natbib}
\usepackage{doi}
\usepackage{amsmath}
\usepackage[]{enumitem}

\title{Surrogate Modeling of Car Drag Coefficient with Depth and Normal Renderings} 

\author{ \href{https://orcid.org/0000-0002-0322-796X}{\includegraphics[scale=0.06]{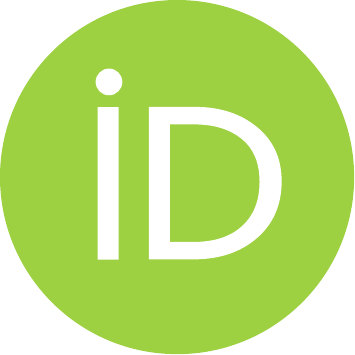}\hspace{1mm}Binyang ~Song}\thanks{Address all correspondence for other issues to this author.}  \\
	Department of Mechanical Engineering\\
	Massachusetts Institute of Technology\\
	Cambridge, MA 02139 \\
	\texttt{binyangs@mit.edu} \\
	\And
	{\includegraphics[scale=0.06]{orcid.pdf}\hspace{1mm}Chenyang ~Yuan} \\
	Toyota Research Institute\\
	Cambridge, MA 02139\\
	\texttt{chenyang.yuan@tri.global} \\
	\AND
	{\includegraphics[scale=0.06]{orcid.pdf}\hspace{1mm}Frank ~Permenter} \\
	Toyota Research Institute\\
	Cambridge, MA 02139\\
	\texttt{frank.permenter@tri.global} \\
	\AND
	{\includegraphics[scale=0.06]{orcid.pdf}\hspace{1mm}Nikos ~Arechiga} \\
	Toyota Research Institute\\
	Los Altos, CA 94022\\
	\texttt{nikos.arechiga@tri.global} \\
	\AND
	{\includegraphics[scale=0.06]{orcid.pdf}\hspace{1mm}Faez ~Ahmed} \\
	Department of Mechanical Engineering\\
	Massachusetts Institute of Technology\\
	Cambridge, MA 02139 \\
	\texttt{faez@mit.edu} \\
}

\renewcommand{\shorttitle}{\textit{arXiv} Surrogate Modeling of Car Drag Coefficient}

\hypersetup{
pdftitle={A template for the arxiv style},
pdfsubject={q-bio.NC, q-bio.QM},
pdfauthor={David S.~Hippocampus, Elias D.~Striatum},
pdfkeywords={First keyword, Second keyword, More},
}

\begin{document}
\maketitle

\begin{abstract}
Generative AI models have made significant progress in automating the creation of 3D shapes, which has the potential to transform car design. In engineering design and optimization, evaluating engineering metrics is crucial. To make generative models performance-aware and enable them to create high-performing designs, surrogate modeling of these metrics is necessary. However, the currently used representations of three-dimensional (3D) shapes either require extensive computational resources to learn or suffer from significant information loss, which impairs their effectiveness in surrogate modeling. To address this issue, we propose a new two-dimensional (2D) representation of 3D shapes. We develop a surrogate drag model based on this representation to verify its effectiveness in predicting 3D car drag. We construct a diverse dataset of 9,070 high-quality 3D car meshes labeled by drag coefficients computed from computational fluid dynamics (CFD) simulations to train our model. Our experiments demonstrate that our model can accurately and efficiently evaluate drag coefficients with an $R^2$ value above 0.84 for various car categories. Moreover, the proposed representation method can be generalized to many other product categories beyond cars. Our model is implemented using deep neural networks, making it compatible with recent AI image generation tools (such as Stable Diffusion) and a significant step towards the automatic generation of drag-optimized car designs. We have made the dataset and code publicly available at \url{https://decode.mit.edu/projects/dragprediction/}.
\end{abstract}

% keywords can be removed

\section{INTRODUCTION}
Engineers often need to work with three-dimensional (3D) representations of an object for design, evaluation, and optimization. At the same time, computer vision researchers have developed powerful deep-learning techniques for various 3D tasks~\cite{Garcia-Garcia2016PointNet:Recognition, Tatarchenko2017OctreeOutputs, Li2018SO-Net:Analysis, Li2022AAutoencoder, Wang2018Pixel2Mesh:Images, Luo2021DiffusionGeneration, Zhou20213DDiffusion, Zeng2022LION:Generation, Nichol2022Point-E:Prompts, Nichol2021GLIDE:Models}, including automatic generation of novel 3D objects. Applying these techniques to design tasks requires evaluating performance metrics at scale. Traditionally, performance evaluation relies on physical simulation, which is time-consuming and computationally expensive. Data-driven surrogate models provide more scalable alternatives. This paper develops a surrogate model for evaluating the aerodynamic drag of 3D vehicles, aiming toward the eventual performance-guided generation of vehicle  designs.

A key challenge in developing a surrogate model is representing shapes in a computationally efficient way that also captures the structure needed to accurately estimate relevant performance metrics. In machine learning, commonly used 3D shape representation methods include voxels, point clouds, and meshes, each affording different advantages and disadvantages. For example, 3D convolutional neural networks (CNNs) are commonly applied to learn structured voxel data~\cite{Prokhorov2010AData, Maturana2015VoxNet:Recognition}, while graph neural networks (GNNs)~\cite{Li2022AAutoencoder, Wang2018Pixel2Mesh:Images} and CNNs generalized to irregular spaces~\cite{Qi2016VolumetricData, Wang2019DominantRecognition, Masci2015GeodesicManifolds} can learn unstructured 3D meshes. In recent years, diffusion models have successfully been leveraged for learning point clouds for 3D shape generation~\cite{Luo2021DiffusionGeneration, Zhou20213DDiffusion, Zeng2022LION:Generation, Nichol2022Point-E:Prompts, Nichol2021GLIDE:Models}. These direct 3D representations are computationally limited to low-resolution shapes, which in turn limits their applications to practical engineering problems.

In addition to direct 3D representations, abstract representations in terms of two-dimensional (2D) renderings have also been explored.
Since technologies for recognizing and generating 2D data is older and more mature than that for learning 3D data, several studies employ 2D renderings or point coordinate matrices to represent 3D shapes~\cite{Ghadai2021Multi-resolutionModels, Su2015Multi-viewRecognition, Achlioptas2017LearningClouds}. Parametric representations are another option to simplify the representation of 3D shapes~\cite{Gunpinar2019ADynamics, Umetani2018LearningDesign, Badias2019AnAnalysis}. These simplified 2D and parametric representations, however, suffer from varying degrees of information loss, and cannot provide sufficient information to reconstruct the corresponding 3D shapes. Accordingly, we propose a new image-based representation of 3D shapes that augments traditional 2D renderings with surface normal and depth information.

We use our representation to train a surrogate model for vehicle drag coefficient prediction, which is a key performance metric that affects not only fuel efficiency but also vehicle aesthetics. As we show, this enables fast and accurate estimation of 3D drag from 2D input. Our contributions are summarized as follows.

\begin{enumerate}
     \item We construct and share a large and diverse set of high-quality car 3D meshes labeled with drag coefficients computed by a fluid dynamics simulation. 
     
     \item We propose a 2D image representation of 3D shapes that annotates 2D renderings with depth and surface normal information using pixel values.
     
     \item We develop a high-performing surrogate model using the proposed representation for car drag coefficient prediction. Leveraging our 2D image representation, we base this model on powerful pre-trained neural networks for image processing tasks.
     %Integrating such a surrogate model into deep generative models can make the generation process performance aware and enable them to generate high-performing car body designs.
     \end{enumerate}
In total, these contributions are a step towards the automatic, performance-aware generation of vehicle body designs. The surrogate model for car drag coefficient prediction also offers an efficient alternative to expensive 3D fluid dynamic simulations. We also hope that our dataset will facilitate the development of various deep-learning techniques for car body design, evaluation, and optimization.

The remainder of this paper is organized as follows. Section 2 provides a detailed review of the relevant literature. In Section 3, we describe our dataset, our novel representation of 3D shapes, and our surrogate model for drag coefficient prediction. Section 4 reports and discusses the effectiveness of the proposed representation and the performance of the surrogate model, and also summarizes the limitations of our approach. 

\section{LITERATURE REVIEW}
The two research areas most relevant to our contributions are 3D object representation and data-driven prediction of drag coefficients.
\subsection{3D Shape Representation and Learning}
In machine learning, 3D shapes are commonly represented as voxels, point clouds, or meshes. Different representations are often matched with different learning algorithms since different algorithms are better suited to exploit the advantages of each representation. For example, similar to CNNs that employ 2D kernels to learn visual features from images, 3D CNNs utilize 3D kernels to capture geometric features from structured 3D spatial data in Euclidean spaces. They are a popular option to learn voxels~\cite{Prokhorov2010AData, Maturana2015VoxNet:Recognition} and occupancy grids for 3D shape recognition~\cite{Garcia-Garcia2016PointNet:Recognition} and generation~\cite{Tatarchenko2017OctreeOutputs}. Since point clouds and meshes are unstructured, prior studies have explored transforming them into regular voxel grids~\cite{Prokhorov2010AData, Wu20153DShapes} or other canonicalized formats~\cite{Wang2018LocalLearning}. However, the sparsity of most 3D data representations makes the computation of the na\"ive 3D convolutional learning challenging. Researchers have proposed a few approaches to mitigate this issue. For example, multiple-resolution 3D CNNs can learn multi-scale features from multi-level voxels~\cite{Boscaini2016LearningNetworks}, while OctNet~\cite{Tatarchenko2017OctreeOutputs} represents its volumetric output as an octree with improved resolutions in the later levels. The voting~\cite{Wang2015VotingDetection} or probing~\cite{Li2016FPNN:Data} schemes in neural networks have been developed to assign varying amounts of computational effort to different regions of sparse data inputs. 

In contrast, a more diverse set of deep learning models have been developed to learn unstructured 3D representations in non-Euclidean spaces (e.g., meshes, manifolds, and point clouds). Inspired by conventional CNNs, a group of researchers developed a variety of CNN variants to learn irregular representations, including localized spectral CNNs~\cite{Qi2016VolumetricData}, anisotropic CNNs~\cite{Wang2019DominantRecognition}, spline-based CNNs~\cite{Fey2017SplineCNN:Kernels}, geodesic CNNs~\cite{Masci2015GeodesicManifolds}, and others. Beyond that, GNNs have been applied to learn both point clouds and meshes for 3D shape recognition~\cite{Li2018SO-Net:Analysis} and generation~\cite{Li2022AAutoencoder, Wang2018Pixel2Mesh:Images}. More recently, diffusion models are becoming an area of active research interest. They have been applied to generate 3D shapes represented by point clouds or similar representations~\cite{Luo2021DiffusionGeneration, Zhou20213DDiffusion, Zeng2022LION:Generation, Nichol2022Point-E:Prompts, Nichol2021GLIDE:Models}. Due to computational cost, the 3D point clouds or meshes generated by these models still present low resolutions, impairing their applications in engineering domains. Prior studies have also explored simple multi-layer perceptrons (MLPs) for mesh texture editing~\cite{Michel2021Text2Mesh:Meshes, Jetchev2021ClipMatrix:Meshes}. 

2D representations have also been explored to represent 3D shapes. A few studies look into representing 3D shapes using 2D images or renderings, which can be processed by standard image learning algorithms~\cite{Ghadai2021Multi-resolutionModels, Su2015Multi-viewRecognition}. Despite the improved computational efficiency, such methods often suffer from information loss. Alternatively, Achlioptas et al.~\cite{Achlioptas2017LearningClouds} proposed a representation that uses the point coordinates of a point cloud as a matrix and trains a generative model with the 2D matrix representation. This approach, however, can only work with point clouds that have a fixed number of points. Another set of studies maps 3D shapes to 2D parameter domains, then trains GANs to generate samples in the 2D domains, and finally converts them to 3D meshes~\cite{Maron2017ConvolutionalCovers, Ben-Hamu2018Multi-chartModeling, Saquil2020Rank3DGAN:Attributes, Alhaija2022XDGAN:Space}. Additionally, implicit representations have also been explored for machine learning tasks. Implicit representations take a latent embedding of a shape and point coordinates as input and assign a value to each point which indicates if this point is inside or outside the shape~\cite{Chen2018LearningModeling, Park2019DeepSDF:Representation}. These representations are often used for 3D shape generation~\cite{Alwala2022Pre-trainReconstruction, Liu2022ISS:Generation}. A group of other studies exploits parametric representations which seek to convey the control points or other prominent features of 3D shapes for machine learning tasks~\cite{Gunpinar2019ADynamics, Umetani2018LearningDesign, Badias2019AnAnalysis}.

In summary, the recognition, evaluation, and generation of 3D shapes using machine learning rely on effective and accurate 3D geometric feature learning. Existing representations of 3D shapes are still greatly limited by their high computational costs, while alternative 2D, implicit, and parametric representations suffer from information loss and may not capture sufficient geometric features for downstream tasks. In this paper, we show that a new representation of 3D shapes using stacked depth and normal renderings is a promising approach, which helps significantly in the downstream task of predicting the drag coefficients of 3D cars.

\subsection{Data-Driven Drag Coefficient Evaluation}
Performance evaluation of 3D shapes is critical in engineering design and optimization. Among them, drag coefficient prediction is critical for car body design. It is traditionally conducted through simulations by solving the nonlinear Navier-Stokes equations for many iterations, which are time-consuming and computationally expensive. The solution methods are too slow to run in conjunction with a generative design or optimization process, which needs to evaluate a large number of candidate designs. To mitigate this issue, researchers have explored combining differentiable partial differential equations (PDEs) solvers with deep learning models to accelerate the simulation results without sacrificing the simulation accuracy significantly~\cite{10.5555/3524938.3525162}. These differentiable PDE solvers often simulate the problem at a coarse resolution and the neural networks are employed to infer the results at higher resolutions. Such an approach speeds up the simulation process but is still too slow to be implemented during the deep generative process. As an alternative to differentiable PDE solvers, data-driven surrogate modeling is a desirable alternative to the simulation approaches in deep learning, and previous work has explored surrogate models for drag coefficient evaluation. 

Parametric representation is commonly used in surrogate modeling of vehicle drag. For instance, Gunpinar et al.~\cite{Gunpinar2019ADynamics} represent a car using the coordinates of a set of control points from the 2D car silhouette and trained computational models to predict its drag coefficient in 2D settings. Their model first reduces the dimension of the representation using principal component analysis and then employs regression models or neural networks to learn the low-dimensional representation for drag coefficient prediction. Likewise, Rosset et al.~\cite{Rosset2023InteractiveFeedback} predicted the pressure field along the car silhouette to optimize 2D car designs. Umetani and Bickel~\cite{Umetani2018LearningDesign} employed a parameterization method to represent simplified cars as vectors that indicate the position of control points and projection heights of the surface points. Then, they learned the representation using regression models, neural networks, or the Gaussian process for drag coefficient prediction. These studies reported that the regression or the Gaussian process models achieved higher explanatory power than the neural network models. Badias et al.~\cite{Badias2019AnAnalysis} used locally linear embeddings to parameterize 3D cars and employed dimensionality reduction and interpolation to predict the drag coefficient of a new car. Limited by their parametric representations, these studies attempt to predict the drag coefficients of simplified cars, such as 2D car silhouettes or 3D cars with mirrors, wheels, and other details removed. This simplification may hinder the applications of such models in practical design contexts. 

Another set of surrogate models learns 2D or 3D car representations to predict drag coefficients. For example, MeshSDF~\cite{Remelli2020MeshSDF:Extraction} learns 3D point clouds obtained from an implicit representation using an irregular CNN (i.e., spline-based CNNs~\cite{Fey2017SplineCNN:Kernels}), which applies to drag coefficient prediction. Similarly, Baque et al.~\cite{Baque2018GeodesicOptimization} exploited a geodesic CNN~\cite{Masci2015GeodesicManifolds} to obtain a latent representation of 3D car meshes for drag coefficient prediction. Another model learns 2D slices of 3D point clouds using regular CNNs~\cite{Jacob2021DeepShapes}, while DEBOSH~\cite{Durasov2021DEBOSH:Optimization} learns meshes using GNNs for the same purpose. Additionally, another class of models obtains the latent representations of 2D~\cite{Thuerey2018DeepFlows} or 3D shapes~\cite{Saha2021ExploitingDesigns} through reconstruction using generative models like variational autoencoders (VAEs) to predict the pressure fields and drag coefficients. Other surrogate models focus on drag prediction of general 3D shapes beyond cars~\cite{Xin2022SurrogateNetworks, TAO2020ApplicationOptimization, SunADesign}. Due to high computational costs, such models can only work with low-resolution 3D representations or simplified 2D representations. This paper focuses on surrogate modeling using the proposed representation of 3D shapes to circumvent the issues of the reviewed approaches.

\section{DATA AND METHOD}
In this section, we detail our main contributions in this paper: A high-quality dataset of 3D car meshes and their drag coefficients computed through computational fluid dynamics (CFD) simulations, a 2D representation generated from 3D car meshes tailored to capturing features important for predicting drag coefficients, and a series of surrogate models trained to predict drag coefficients from the 2D representation as a regression task~\footnote{The dataset and the surrogate models introduced in this paper can be found: Github link}. 

\subsection{Car Data and CFD Simulation}
First, we detail our 3D car dataset and CFD simulations for obtaining drag coefficients from 3D mesh data.
\subsubsection{Car Data}
The 3D car meshes used in this paper are initially from the ShapeNet V1 dataset~\cite{Chang2015ShapeNet:Repository}, which contains 7,497 3D car meshes with varying surface qualities. A substantial percentage of the original car meshes from ShapeNet are not watertight, with unsealed areas or holes on the surfaces. We need high-surface-quality car meshes in order to achieve reliable CFD simulation results when computing car drag coefficients. Therefore, we manually checked the surface quality of each car mesh from ShapeNet and selected a subset of 2,474 high-quality car meshes. Since most of the selected meshes are still imperfect, we further repaired them using the repair module in Autodesk Netfabb Premium. It should be noted that this dataset covers a variety of car configurations, such as pick-up trucks, sedans, sport utility vehicles, wagons, and combat vehicles. The diversity helps our learned surrogate models generalize across all cars.

%\begin{figure}[htbp]
%    \centering
%    \includegraphics[width=0.9\columnwidth]{figure/samples.pdf}
%   \caption{Car samples from our dataset.}
%    \label{fig:sample}
%\end{figure}

In addition, we employed two different approaches to augment the original dataset. First, we resized the width of each car using a random coefficient between 0.83 (i.e., $1/1.2$) to 1.2. The resizing augmentation created another 2,474 cars with slightly different widths and drag coefficients from the original cars, resulting in a dataset of 4,958 different cars in total. Second, since the car meshes are not perfectly bilaterally symmetric but their drag coefficients are invariant to bilateral flipping, we employed a flipping augmentation to create another 4,948 cars, which have exactly the same drag coefficients as the cars without this augmentation. After the augmentations, we obtain a dataset of 9,896 cars. To avoid data leakage, we only treat the 2,474 unique cars from the original dataset as independent samples when splitting the dataset to train the surrogate model. For every car in any of the training, validation, or test sets, all of its resized and flipped versions belong to the same set.

\subsubsection{CFD Simulation}
\begin{figure}[htbp]
    \centering
    \includegraphics[width=9cm]{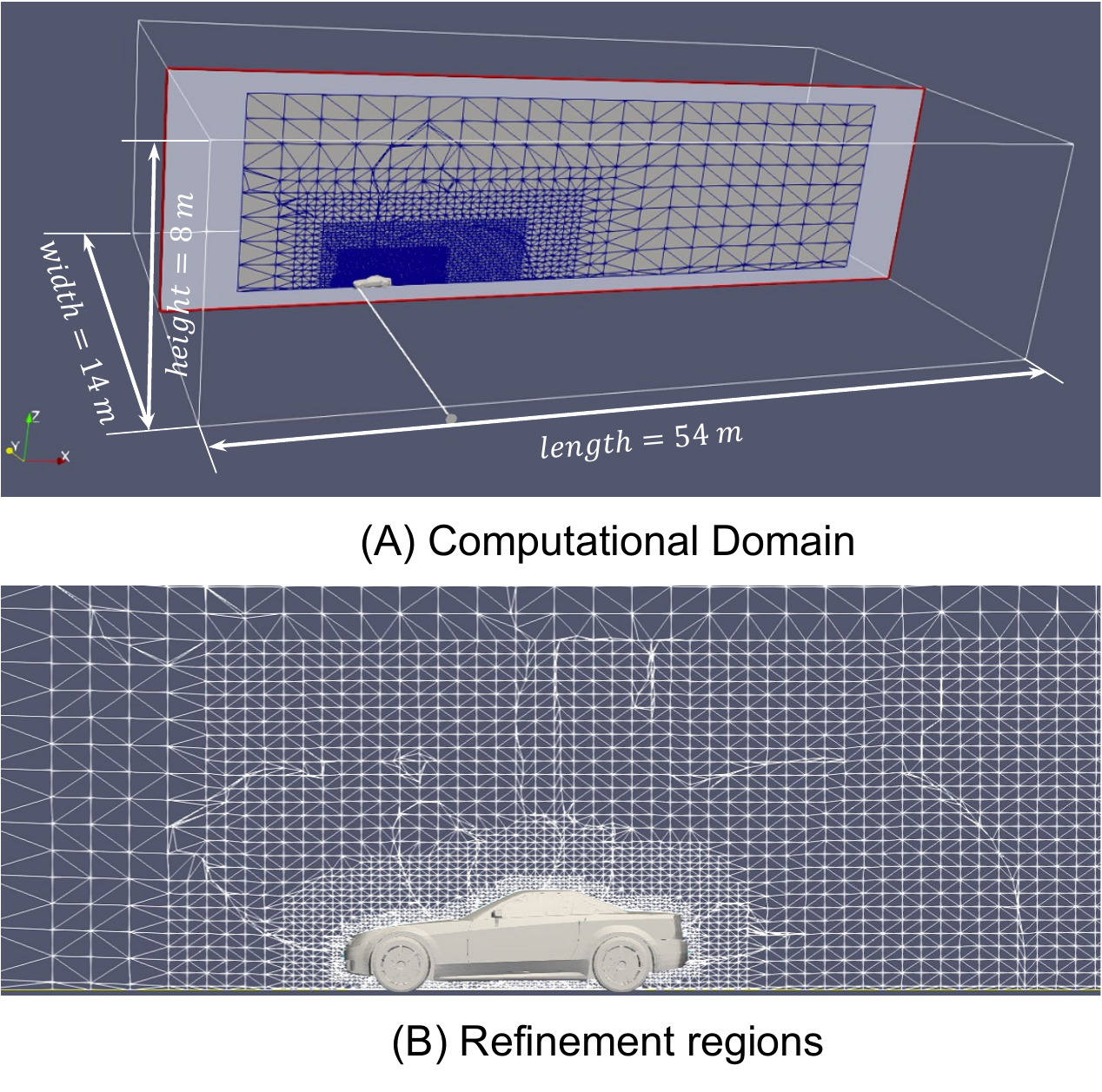}
    \caption{The computational domain for the CFD simulation}
    \label{fig:cfd}
\end{figure}

The drag coefficient of each car is computed by a CFD simulation using OpenFOAM. During mesh preparation, all cars are normalized to have the same length of 3.5 meters to ensure the defined computational domain is suitable for all cars. The computational domain for simulation is then created, serving as a virtual wind tunnel to simulate the airflow around a car, as shown in~\ref{fig:cfd}-A. The height, width, and length of the virtual tunnel are 8 meters, 14 meters, and 54 meters, respectively. In order to simulate flow dynamics around the car body more accurately, the computational domain is refined to a smaller mesh size, which becomes coarse away from the car surface, as shown in~\ref{fig:cfd}-B. This meshing strategy is applied to all car configurations (e.g., sedans, sports utility vehicles, combat cars, and pick-up trucks) in our dataset.

On this basis, the inlet velocity and turbulence parameters are set as the inlet conditions, while outlet pressure is specified as the outlet condition. The car surface and road are set to be stationary walls. The sides and top of the computational domain are specified as symmetry boundaries. The steady-state ``SimpleFoam'' solver and the fluid flow ``PotentialFoam'' solvers are selected for the simulation. The primary boundary conditions and solver settings are listed in Tables~\ref{tab:boundary} and~\ref{tab:solver}. During the simulation, 300 iterations were conducted for each car, which can achieve the required accuracy for concept-level studies~\cite{Biswas2019DevelopmentTool}. Since the drag coefficient outputs may fluctuate during the simulation process, we use the average value from the last 50 iterations as the final output from the simulation.

\begin{table}[t]
\caption[Table]{Boundary conditions}\label{tab:boundary}
\centering{%
\begin{tabular}{ll}
\toprule
Tunnel inlet & Velocity inlet, velocity = 40km/h \\
Tunnel outlet & Pressure outlet \\
Tunnel sides & Symmetry \\
Tunnel top & Symmetry \\
Tunnel road & No slip wall, with prism layer \\
Car body & No slip wall \\
\bottomrule
\end{tabular}
}
\end{table}

\begin{table}[t]
\caption[Table]{Solver settings}\label{tab:solver}
\centering{%
\begin{tabular}{ll}
\toprule
Gradient scheme & Linear \\
Divergence scheme (momentum) & Linear upwind \\
Divergence scheme (turbulence) & Upwind \\
Laplacian scheme & Linear \\
Interpolation scheme & Linear \\
Pressure solver & GMAG \\
Velocity solver & Smooth solver \\
No of Non-orthogonal corrections & 2 \\
\bottomrule
\end{tabular}
}
\end{table}

\subsection{2D Representation of 3D Shapes}
In prior work, voxels, point clouds, and meshes are commonly used to represent 3D shapes. They each require different deep neural networks to learn and rely on intensive computational resources to capture fine-grained, high-resolution 3D features. For car body design, we only focus on the surface of the car and ignore any interior architecture. In this paper, we aim to propose a more information-efficient method to represent 3D shapes like car bodies, which supports learning 3D information more effectively and affordably for drag coefficient prediction.

\begin{figure}[htbp]
    \centering
    \includegraphics[width=9cm]{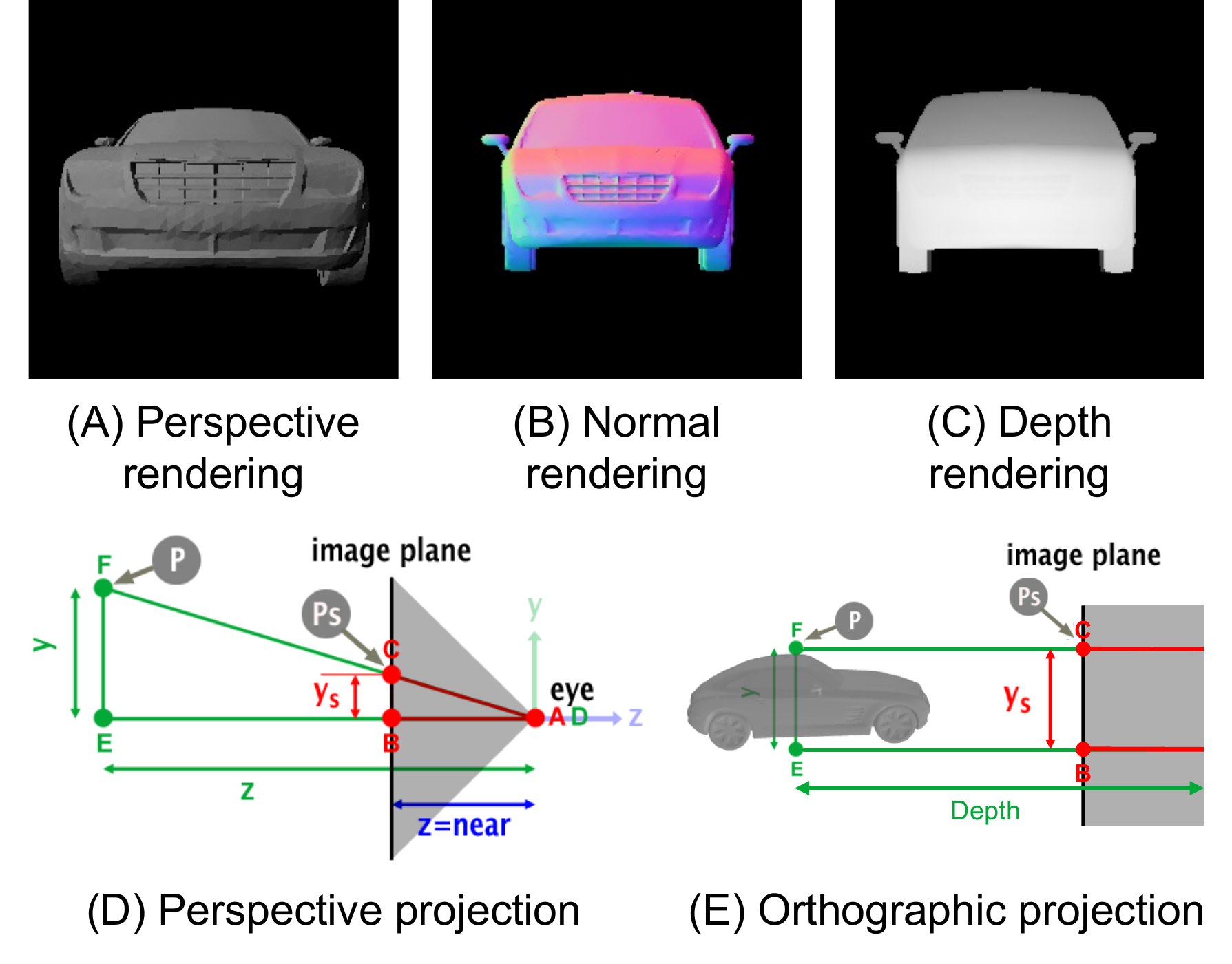}
    \caption{The proposed 2D representation of 3D shapes}
    \label{fig:new representation}
\end{figure}

Since machine learning methods for 2D data learning are more explored than those for 3D data learning, 2D renderings have become an option to represent 3D shapes in many studies. However, the commonly used perspective 2D renderings~\ref{fig:new representation}-A are generated through perspective projection~\ref{fig:new representation}-D, which causes geometric distortion and information loss for machine learning. Accordingly, we propose a new 2D representation of 3D shapes that consists of two types of renderings, namely the normal rendering~\ref{fig:new representation}-B and the depth rendering~\ref{fig:new representation}-C, generated through orthographic projection~\ref{fig:new representation}-E. The points facing the cameras are first projected to the image space through a projection defined by Eq.~\ref{eq:proj}. Herein, $P_{\text{camera}}$ and $P_{\text{world}}$ represent point coordinates (i.e., x, y) in the rendering and real-world space, respectively. $Scale_{\text{x}}$ and $Scale_{\text{y}}$ denote the scaling factors that are determined by the position and angle of the camera and the size of the rendering. Specifically, the pixel values of the normal rendering encode the unit normal vector at each point of the mesh, with the $x~(\text{Norm}_{x})$, $y~(\text{Norm}_{y})$, and $z~(\text{Norm}_{z})$ coordinates mapped to the red ($\text{Color}_{R}$), green ($\text{Color}_{G}$), and blue ($\text{Color}_{B}$) color channels, respectively, as shown by Eq.~\ref{eq:norm}. The pixel values of the depth rendering encode the depth of each point, i.e., the distance ($\text{Dist}$) between the camera and the point, as formulated by Eq.~\ref{eq:depth}.

\begin{equation}
\begin{aligned}
P_{\text{camera}} = P_{\text{world}} \times \begin{bmatrix}
Scale_{\text{x}} & 0\\
0 & Scale_{\text{y}}
\end{bmatrix}, \\
\label{eq:proj}
\end{aligned}
\end{equation}

\begin{equation}
\begin{aligned}
\text{Color}_{R} = \text{Norm}_{x}, ~\text{Color}_{G} = \text{Norm}_{y}, ~\text{Color}_{B} = \text{Norm}_{z}, \\
\label{eq:norm}
\end{aligned}
\end{equation}

\begin{equation}
\begin{aligned}
\text{Color}_{R} = \text{Color}_{G} = \text{Color}_{B} = \text{Dist}. \\
\label{eq:depth}
\end{aligned}
\end{equation}

\begin{figure}[htbp]
    \centering
    \includegraphics[width=9cm]{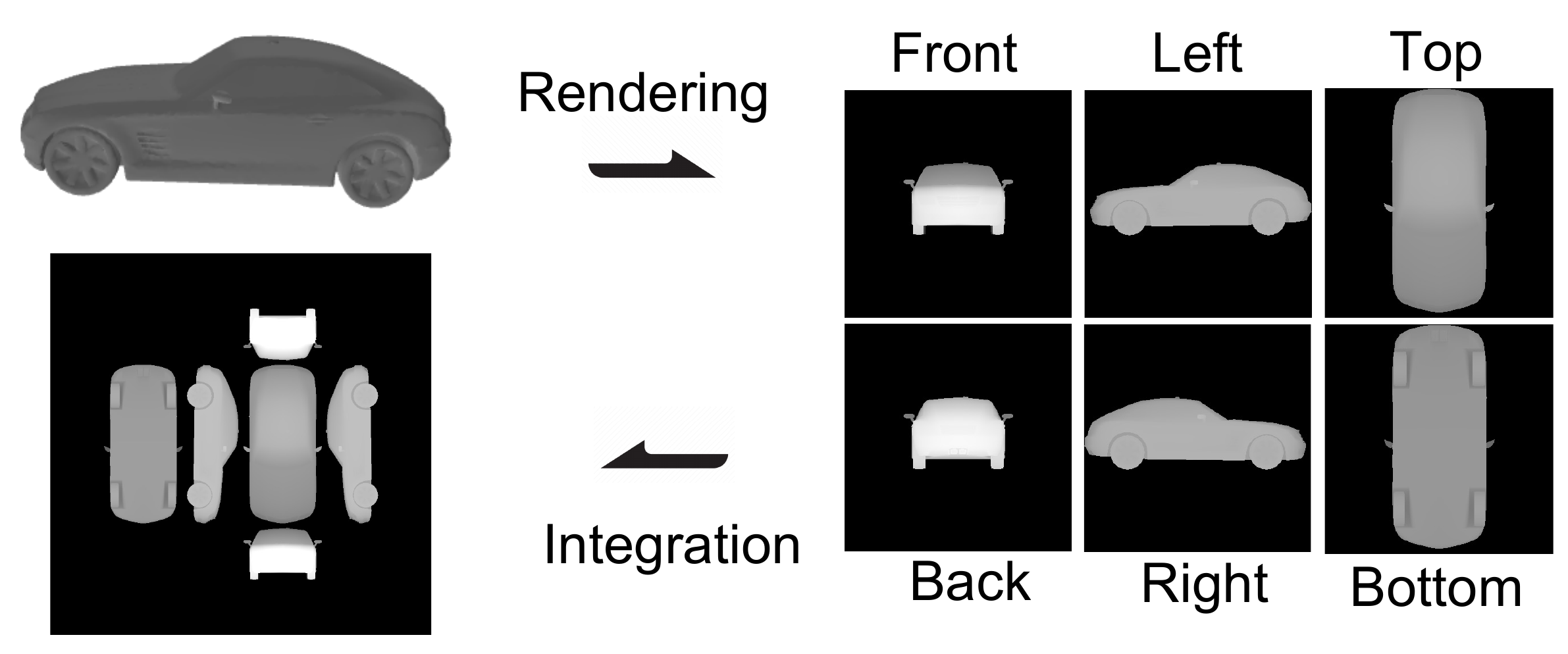}
    \caption{The conversion process from a 3D mesh to 2D renderings, and to an integrated representation}
    \label{fig:rendering conversion}
\end{figure}

According to the definition, the depth and normal renderings capture the point-wise positional and surface information of 3D shapes respectively. In order to capture the geometric features of a car comprehensively, we generate the normal and depth renderings from six orthographic views: front, rear, top, bottom, left, and right. Then, the six single-view renderings are integrated into a single image. With the combined information from all six single-view renderings, the integrated 2D representation conveys 3D geometric information and be potentially converted back to corresponding 3D shapes. Figure~\ref{fig:rendering conversion} describes the process using the depth rendering of a car. Building on the render module of the kaolin python package developed by NVIDIA\footnote{https://github.com/NVIDIAGameWorks/kaolin.}, we develop a differentiable render for 3D to 2D rendering and a separate module for six view integration to produce the 2D representation for each car. The integrated normal and depth renderings are used as the 2D representation of 3D shapes in this paper. We verify the effectiveness of our proposed representation by developing surrogate models to predict car drag coefficients from our 2D representation.

\subsection{Surrogate Model}
Our proposed 2D representation enables us to represent 3D shapes using 2D pixel data. We next develop and compare three surrogate models that take the 2D representation of a car as input and predict its drag coefficient. In this paper, the 2D representations of all cars in our dataset are images with 3 color channels and a dimension of $384\times384$, which is the input to all of our surrogate models.

We explore both the CNN-based and transformer-based computer vision models to learn features from the 2D representation of cars. In a set of pilot experiments, we first compare a few different pre-trained CNN-based models, including InceptionV3~\cite{Szegedy2016RethinkingVision}, ResNet~\cite{He2016DeepRecognition}, and ResNeXt~\cite{Xie2016AggregatedNetworks}. In general, they perform similarly after careful hyper-parameter tuning, and ResNeXt is selected in our study because it performs slightly better than the others. The proposed representation integrates six single-view renderings, which exhibit correspondence and convey complementary information for drag coefficient prediction. This characteristic of the representation motivates us to involve attention mechanisms in the surrogate model. Furthermore, since transformer-based image models can capture the interactions between different image regions through the embedded self-attention mechanism, we also compare the CNN-based models against one transformer-based model, the vision transformer (ViT)~\cite{Dosovitskiy2020AnScale}.

\begin{figure}[htbp]
    \centering
    \includegraphics[width=8cm]{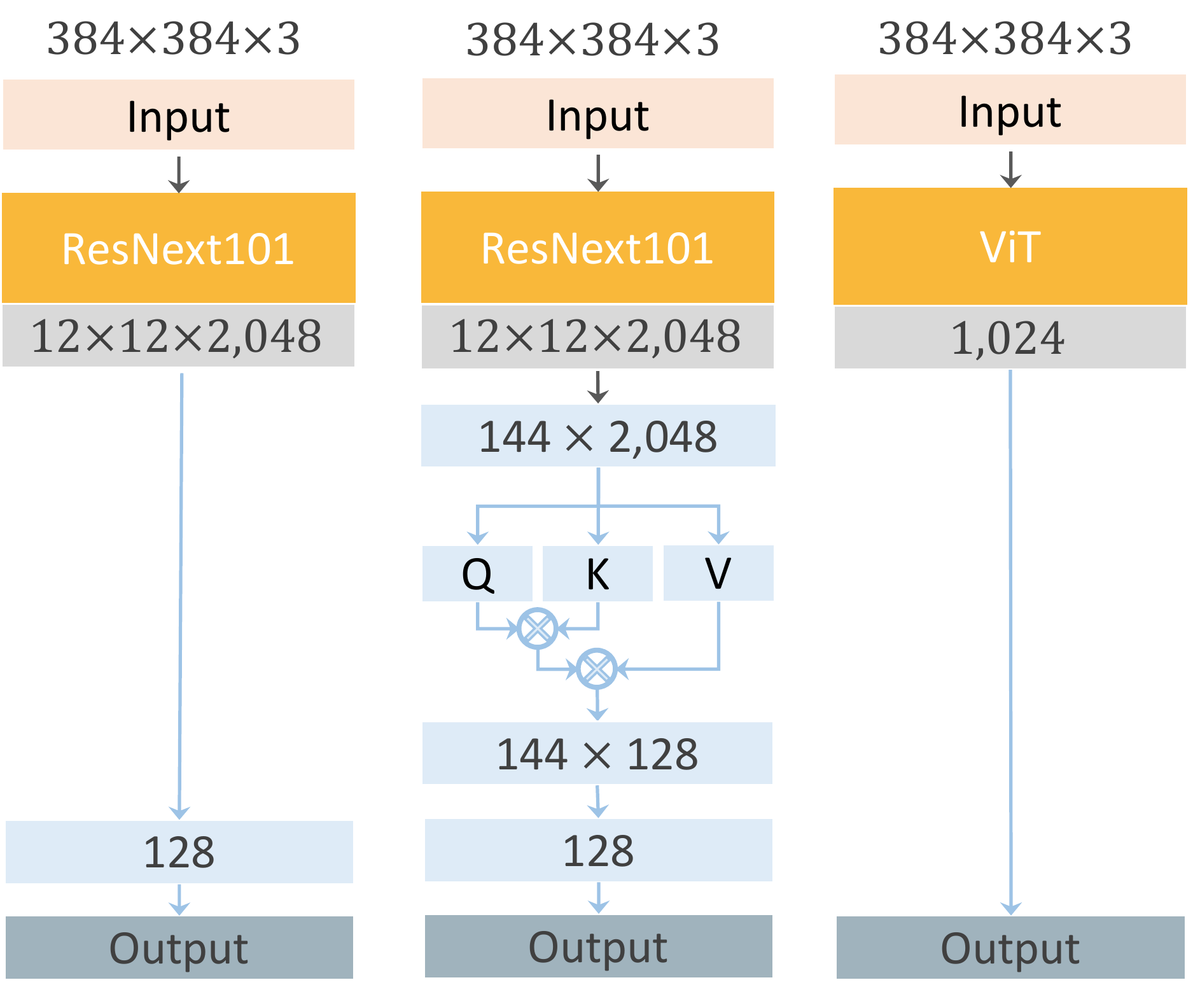}
    \caption{The architectures of the three surrogate models using different embedding modules or different attention mechanisms}
    \label{fig:three}
\end{figure}

The first model (Figure~\ref{fig:three}-A) employs the pre-trained ResNeXt ``$101_32\times8d$" module to embed the image input. The output from the ResNeXt embedding module exhibits a dimension of $12\times12\times2,048$, which is flattened. Following that, a linear layer with 128 neurons is attached before the output layer. We name this model ``ResNeXt" in this paper.

The second model (Figure~\ref{fig:three}-B) applies a self-attention mechanism to enhance the learning of the interactions between different image regions. Specifically, it reshapes the output from the ResNeXt embedding module to $144\times2,048$, which is seen as a set of 144 latent features with a dimension of 2,048. A self-attention mechanism with a latent dimension of $128$ is applied to capture the interactions between the image regions. Then, the output from the self-attention mechanism is flattened and projected to a lower dimension ($128$) through a linear layer as the final embedding to predict the car drag coefficient. This model is referred to as ``attn-ResNeXt" hereafter.

The third model (Figure~\ref{fig:three}-C) utilizes a pre-trained ViT module to embed the image input. We compare two different-sized ViT models, including the ``vit-large-patch32-384" model and the ``vit-base-patch16-224" model, and achieve slightly better performance from the former. Accordingly, ``vit-large-patch32-384" was selected for building the third surrogate model. The pooled output from the transformer embedding module is used as the final embedding to predict the car drag coefficient. We call this model ``ViT" in this paper.

\begin{figure}[htbp]
    \centering
    \includegraphics[width=11.5cm]{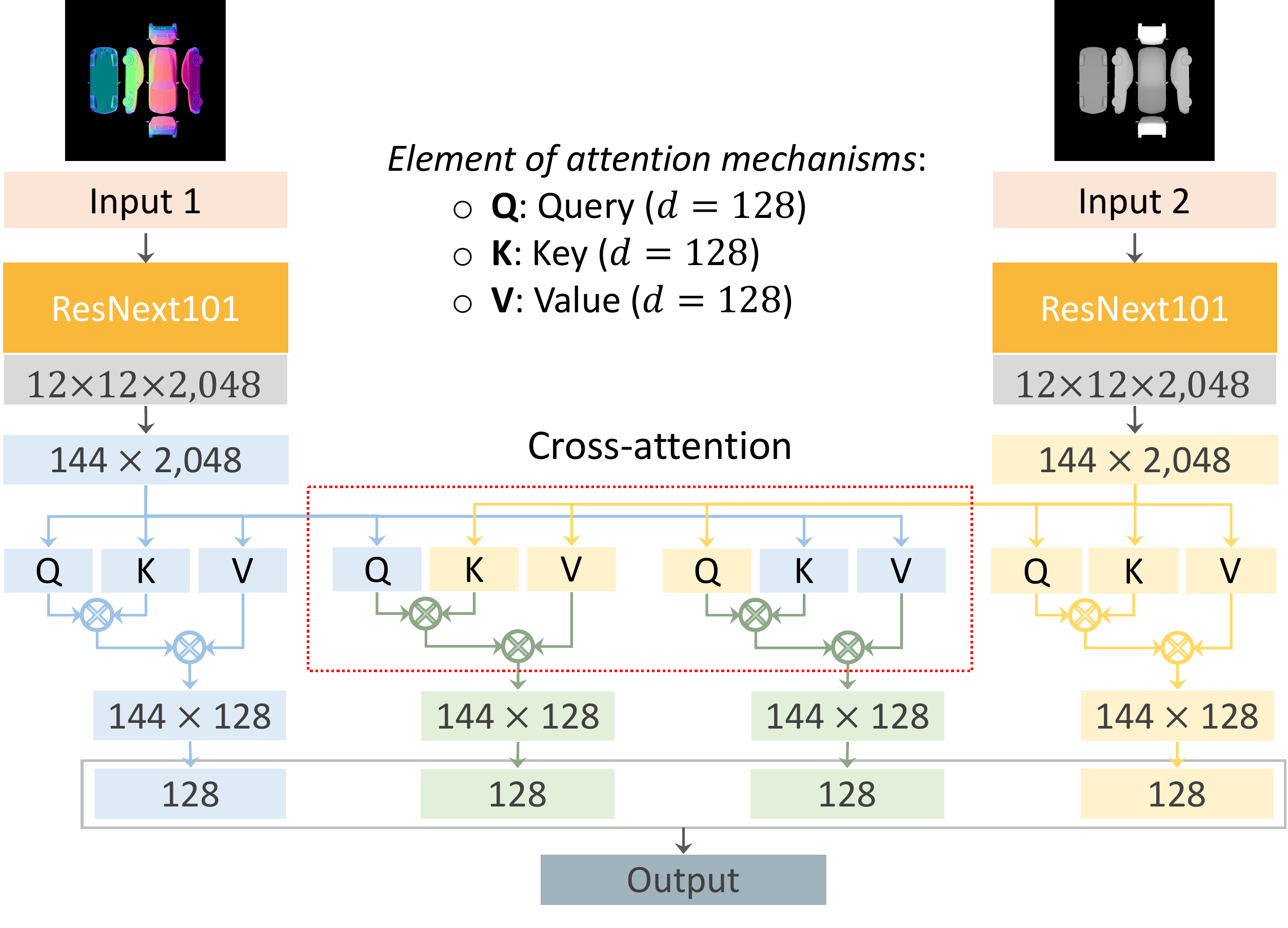}
    \caption{The surrogate model fusing features of both the normal and depth renderings using a symmetric cross-attention mechanism}
    \label{fig:resnext}
\end{figure}

Since the surrogate models introduced above can learn from only one of the normal/depth renderings at a time, we further explore if fusing the features of the normal and depth renderings can improve the prediction performance. After fine-tuning the hyperparameters of all three surrogate models, we select the best among the three for this exploration, which is the attn-ResNeXt model in this study. Specifically, we fuse two attn-ResNeXt models respectively pre-trained on the normal and depth renderings using a symmetric cross-attention mechanism, as shown in Figure~\ref{fig:resnext}. The cross-attention mechanism is expected to capture the interactions between the regions respectively from the normal and depth renderings. Then, the outputs from the self-attention and cross-attention mechanisms are flattened and projected to a lower dimension ($128$) through linear layers, which are then concatenated as the final embedding to predict the car drag coefficient. During training, the fused model is initialized with the pre-trained weights from both the normal rendering model and the depth rendering model to transfer the knowledge learned from the single types of renderings to the fused model. This approach has been proven beneficial for avoiding modality failure~\cite{Du2021ImprovingTeachers, Song2023ATTENTION-ENHANCEDEVALUATIONS}. We refer to this model as ``fused" hereafter.

The hyperparameters of these surrogate models are determined through a set of pilot experiments. In the experiments, all the trainable parameters are unfrozen. The pre-trained ResNeXt and ViT image embedding modules are fine-tuned on our data. We split the entire dataset into the training, validation, and test sets following a ratio of 0.7:0.15:0.15. All models are trained on the same training-validation-test split for easy comparisons. We employ different learning rates ranging from $2\times10^{- \ 5}$ to $8\times10^{- \ 5}$ to train different models with different image inputs. We also apply a decay of $0.96$ to schedule the learning rate during the training process. We end the training process if the validation loss does not decrease for 20 consecutive epochs.

\section{RESULTS AND DISCUSSION}
This section describes our CFD simulation results and compares the performances of different surrogate models based on the proposed 2D representation. To evaluate the models, we report the coefficient of determination ($R^2$ value) and the mean squared prediction error (MSE). To illustrate sensitivity to initialization, we train each model five times and report the average values of these metrics. We also compare our best surrogate model against two baseline models from prior studies.

\subsection{CFD Simulation Results}
As described in the last section, our dataset originates from 4,948 car meshes obtained from ShapeNet.
Drag coefficients were successfully simulated for $4,535$ of these meshes using OpenFOAM. To increase the size of the dataset, we flip each car left to right (which leaves the drag coefficient unchanged), giving a total of $4,535 \times 2 = 9,070 $ training examples.

\begin{figure}[htbp]
    \centering
    \includegraphics[width=10cm]{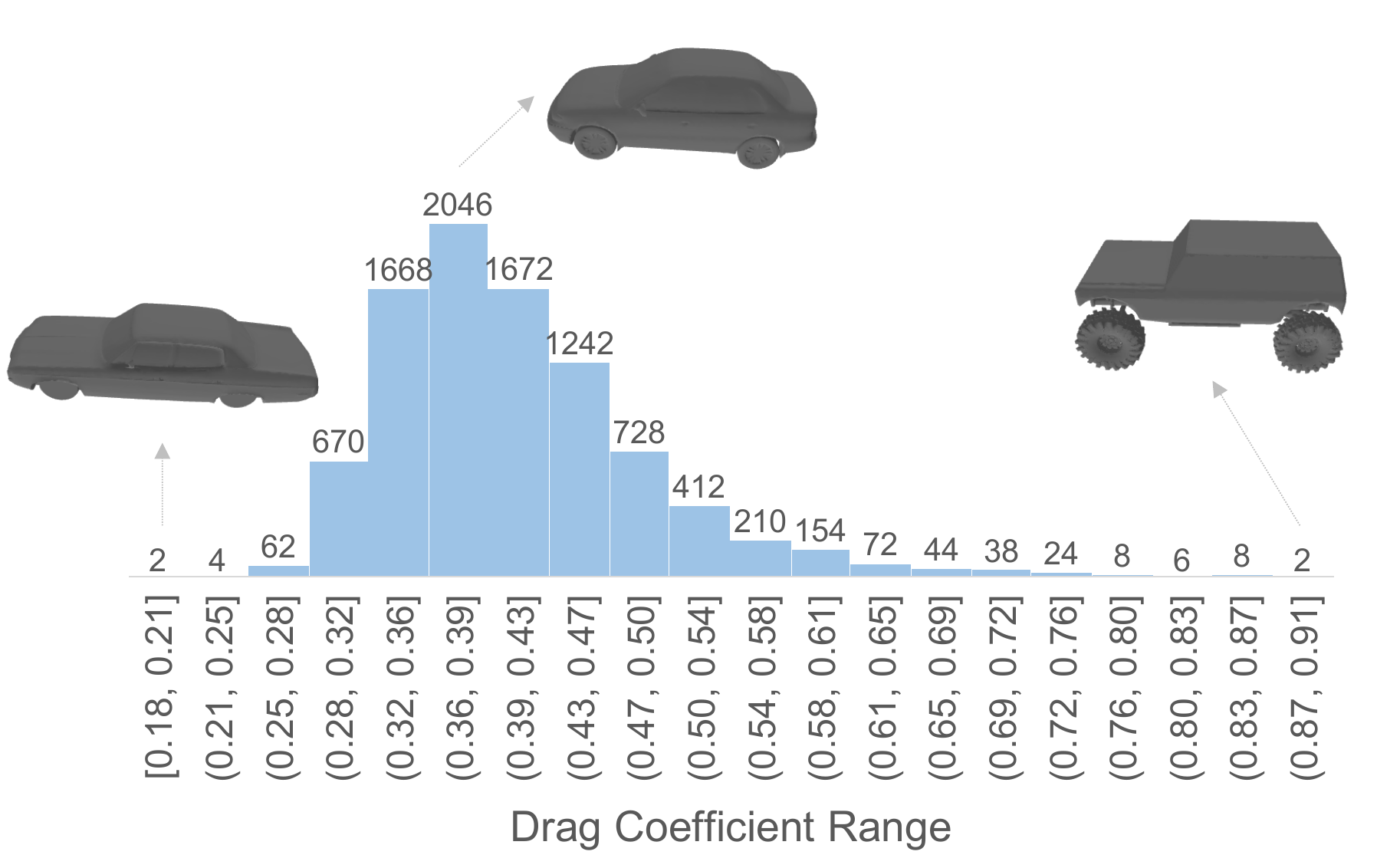}
    \caption{The distribution of the drag coefficients and three example cars from the lowest, biggest, and highest drag coefficient categories, respectively}
    \label{fig:distribution}
\end{figure}

The computed drag coefficients range from 0.175 to 0.907. Figure~\ref{fig:distribution} shows their distribution and three sample vehicle images from different drag coefficient regimes. The data is concentrated on the interval [0.28, 0.65].

\subsection{Performance of Different Surrogate Models}
\begin{figure}[htbp]
    \centering
    \includegraphics[width=13cm]{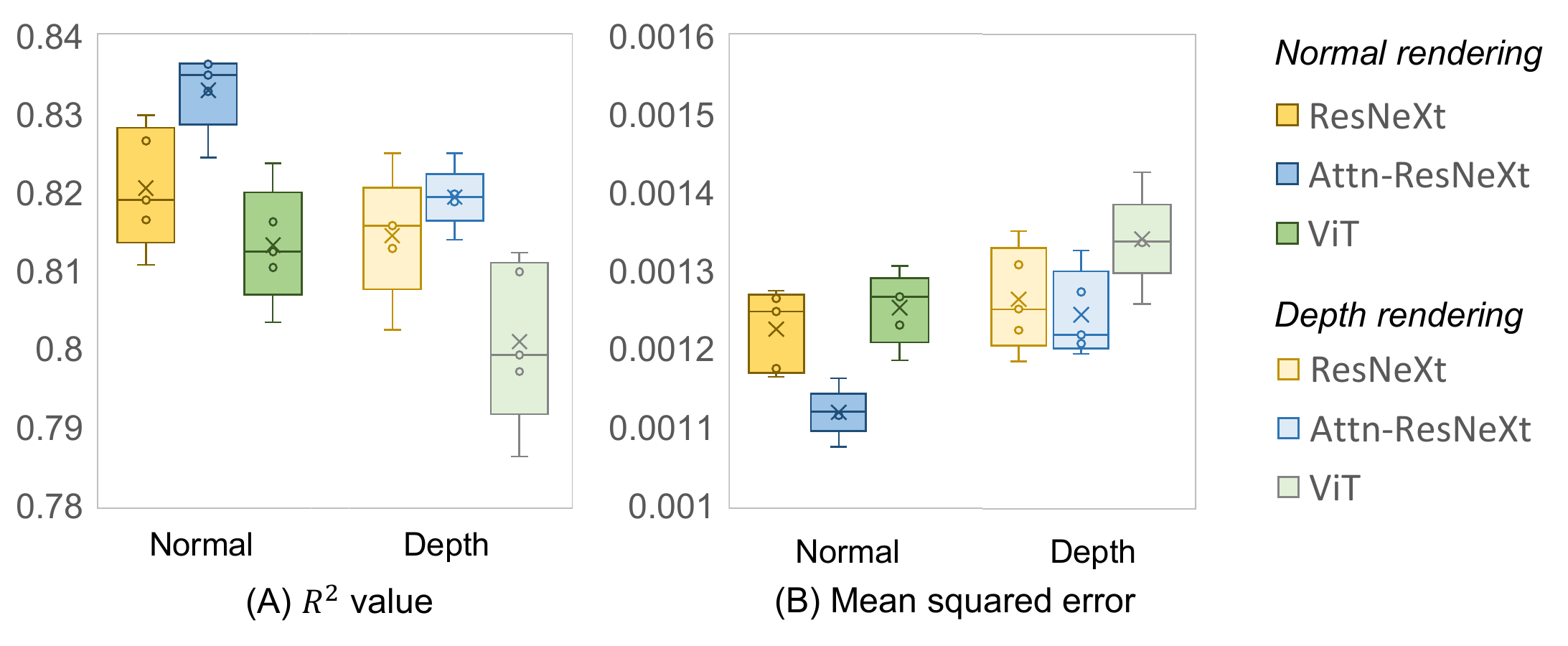}
    \caption{The performance comparison among the three surrogate models using different rendering inputs}
    \label{fig:surrogate}
\end{figure}

We first compare the drag coefficient prediction of six different surrogate models. Each model employs one of the three architectures depicted in Figure~\ref{fig:three} and is trained on either depth or surface normal renderings. Figure~\ref{fig:surrogate} illustrates the performance of each model. Among the three architectures, the attn-ResNeXt model achieves the highest $R^2$ values and the lowest MSE values. The comparison between ResNeXt and attn-ResNeXt suggests that the self-attention mechanism improves the fusion of information from different image regions. Both ResNeXt and attn-ResNeXt outperform the ViT model. A possible reason is that ResNeXt contains far fewer trainable parameters than the ViT model (about 86 million vs about 2 billion) and overfits our relatively small dataset to a lesser degree. 

\begin{figure}[htbp]
    \centering
    \includegraphics[width=11cm]{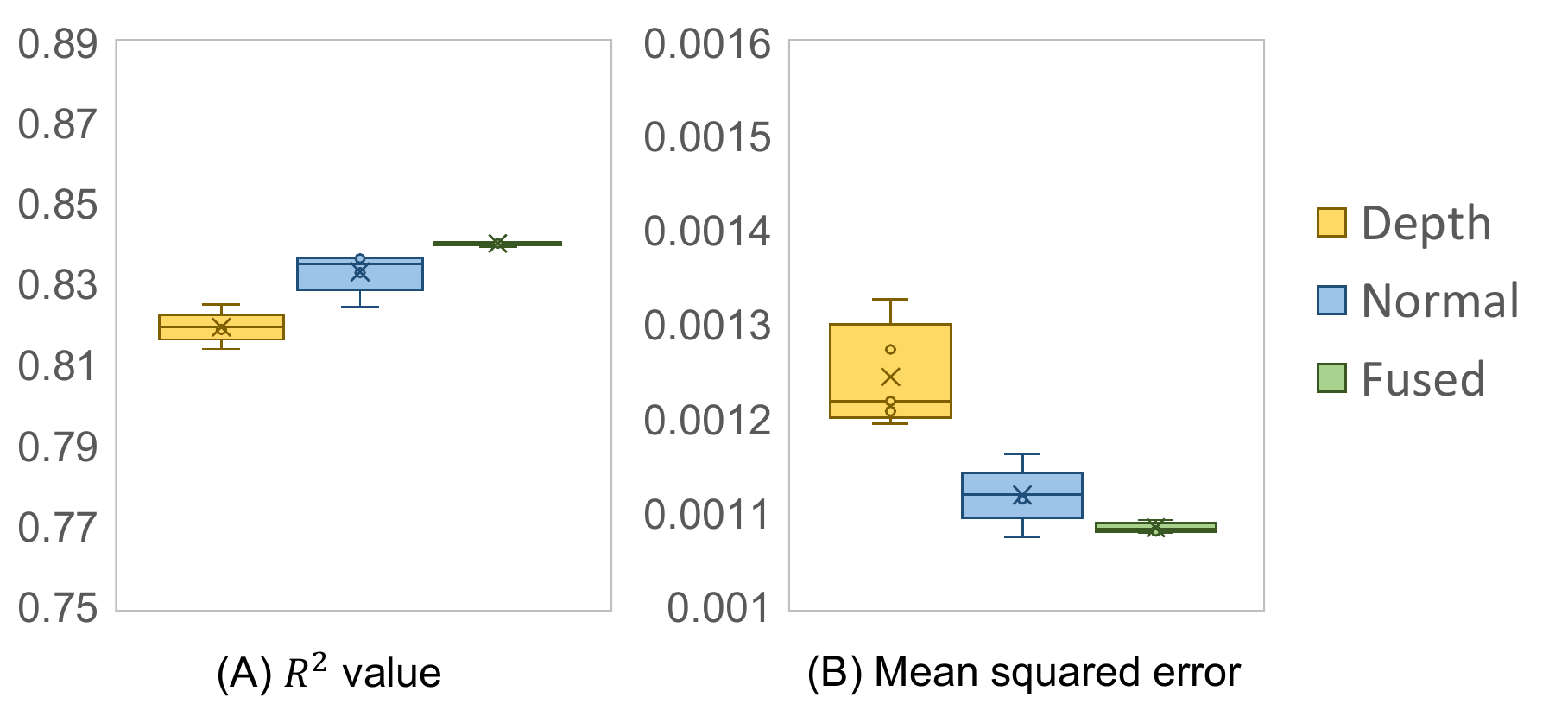}
    \caption{The performance comparison among the two attn-ResNeXt models respectively using the normal and depth renderings and the fused model using both renderings}
    \label{fig:fuse}
\end{figure}

We next illustrate that combining the normal and depth information enhances the performance of the surrogate model.  
We fuse these features using a symmetric cross-attention mechanism as depicted in Figure~\ref{fig:resnext}. Moreover,
we train this fused model using the \emph{transfer learning} paradigm; that is, we initialize the training of the fused model using the weights of the attn-ResNeXt models respectively pre-trained on the normal and depth renderings. Figure~\ref{fig:fuse} illustrates the superior performance of the fused model, and significantly reduced sensitivity to initialization of the training procedure, as indicated by the variance of the $R^2$ values and MSE values.  

\begin{figure}[htbp]
    \centering
    \includegraphics[width=9cm]{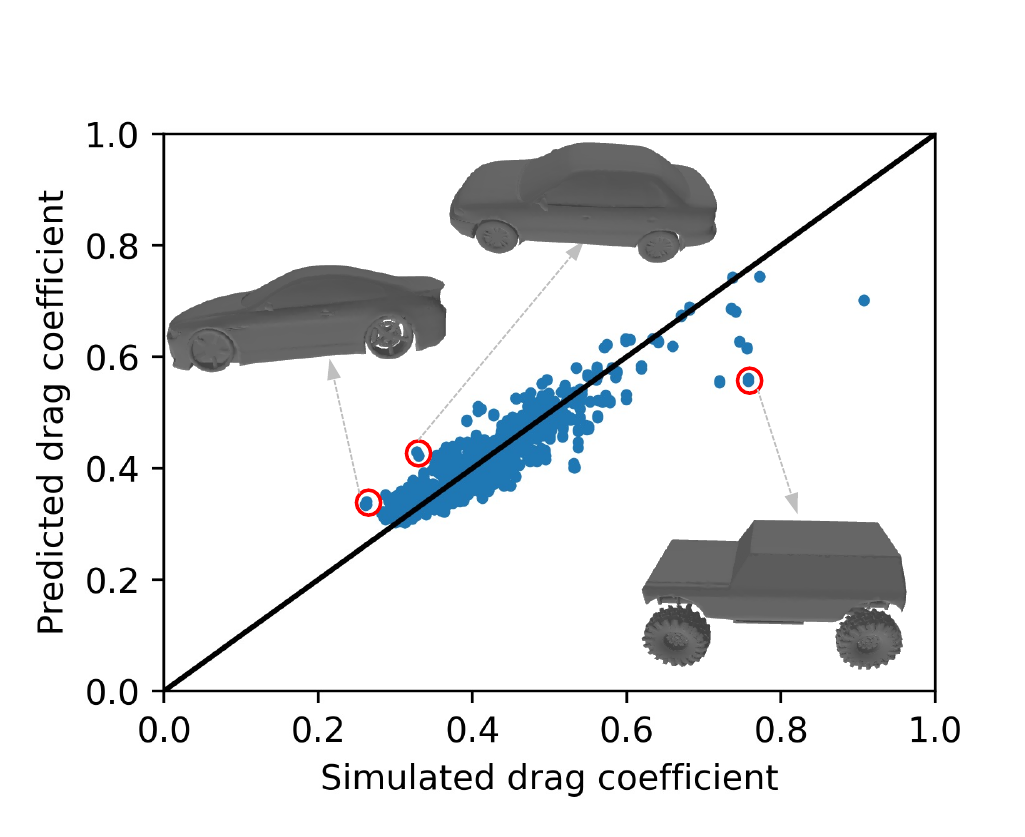}
    \caption{The comparison between predicted and ground-truth values}
    \label{fig:prediction}
\end{figure}

Figure~\ref{fig:prediction} illustrates how the accuracy of the model depends on the ground-truth drag coefficient.  As indicated, the prediction exhibits increasing deviations in the lowest and highest drag coefficient ranges. One major reason is that we have much fewer car samples with very low or high drag coefficients in our dataset (Figure~\ref{fig:distribution}). Accordingly, the model exhibits higher average prediction errors in the lowest and highest drag coefficient ranges, as listed in Table~\ref{tab:error}.

Evaluation of the surrogate models is also significantly faster than drag coefficient computation via CFD simulation. Indeed, it takes in total 20 seconds to evaluate the drag coefficients for 1,362 cars using an NVIDIA RTX A5000 GPU. In comparison, the CFD simulation of a single car takes about 6 minutes on average using a Lambda computer with 12 Intel Xeon(R) E5-1650 CPUs. Finally, the surrogate models are also auto-differentiable and hence more easily incorporated into optimization routines.

\begin{table}[t]
\caption[Table]{The variation of the prediction error with the simulated drag coefficient}
\label{tab:error}
\centering{%
\begin{tabular}{ll}
\toprule
Drag Coefficient Range & Average Prediction Error\\
\hline
[0.18,0.3] & 0.032\\
(0.3, 0.4] & 0.021\\
(0.4, 0.5] & 0.023\\
(0.5, 0.6] & 0.029\\
(0.6, 0.7] & 0.021\\
(0.7, 0.8] & 0.092\\
(0.8,0.91] & 0.218\\
\bottomrule
\end{tabular}
}
\end{table}

\subsection{Effectiveness of the Proposed Representation}
In this subsection, we verify the effectiveness of the proposed representation by comparing its informativeness with single-view renderings and the perspective renderings. Beyond that, we also compare the performance of our surrogate model with the baseline models from two prior studies. The best surrogate model identified in the last subsection, attn-ResNeXt, is used for the following experiments.

\begin{figure}[htbp]
    \centering
    \includegraphics[width=11cm]{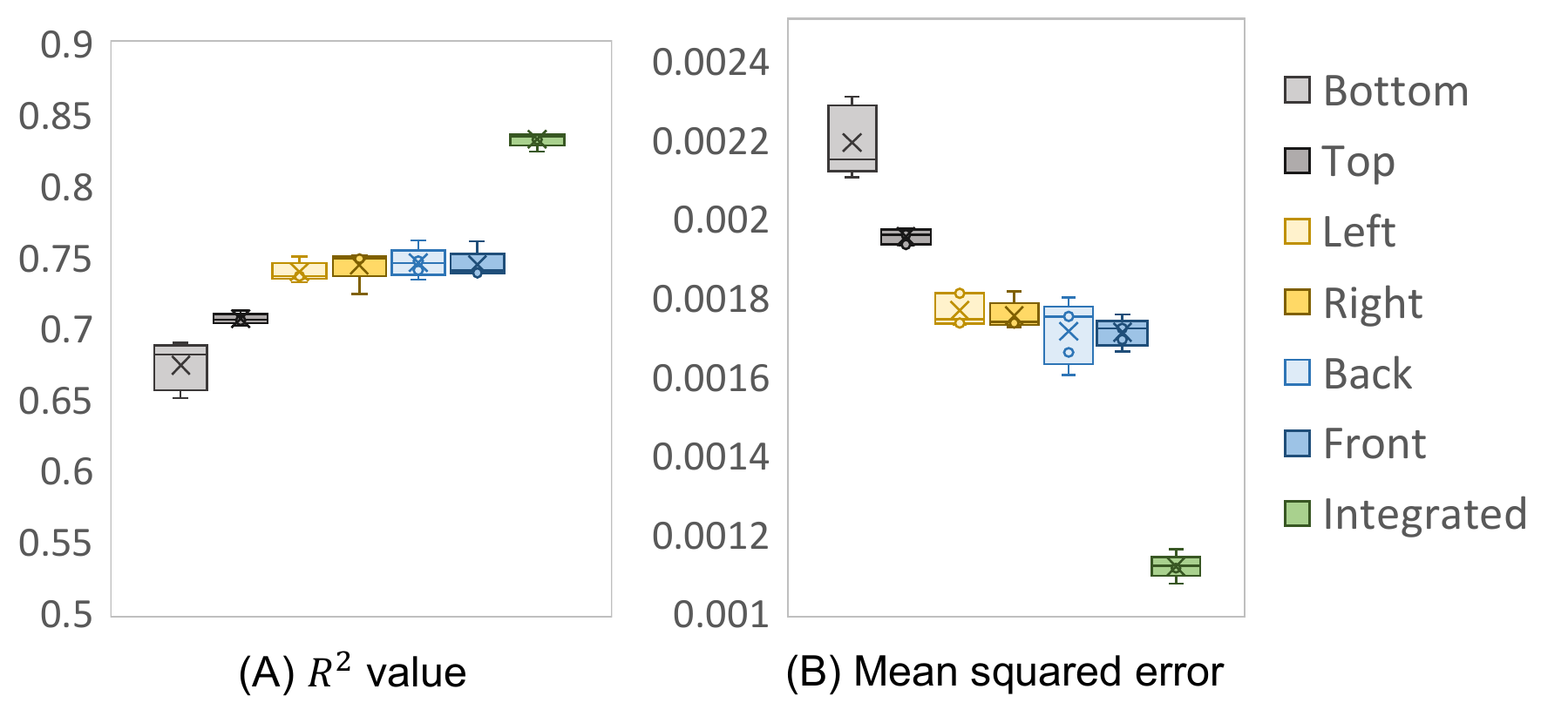}
    \caption{The performance of the surrogate models using the single-view normal renderings and the integrated normal renderings, respectively}
    \label{fig:views}
\end{figure}

Compared to the single-view renderings, the integrated rendering is more informative for car drag coefficient evaluation. In this set of experiments, the attn-ResNeXt model takes the single-view normal renderings and the integrated normal renderings as input, respectively. Figure~\ref{fig:views} depicts their $R^2$ and MSE values. The model taking the integrated renderings as input exhibits the highest $R^2$ value and lowest MSE value compared to all other models taking the single views as input. It is intuitive that the integrated renderings contain the geometric information of a car more comprehensively than any single-view rendering. 

Among all single-view renderings, the front, back, left, and right views provide similar amounts of information for drag coefficient evaluation, leading to similar $R^2$ and MSE values. The bottom view is least informative for this task. In car body design, the streamlined design and the frontal area of a car affect the car's drag coefficient significantly. Every single view contains part of the information. For example, the front and back views reflect the frontal area and the front or rear part of the streamlined design, while the left and right views show the entire streamlined design from two directions. The top view describes the top half of the streamlined design, which is often more informative than the bottom half depicted by the bottom view. The amount of relevant information conveyed by each view greatly determines the explanatory power of the corresponding model.

\begin{figure}[htbp]
    \centering
    \includegraphics[width=11.5cm]{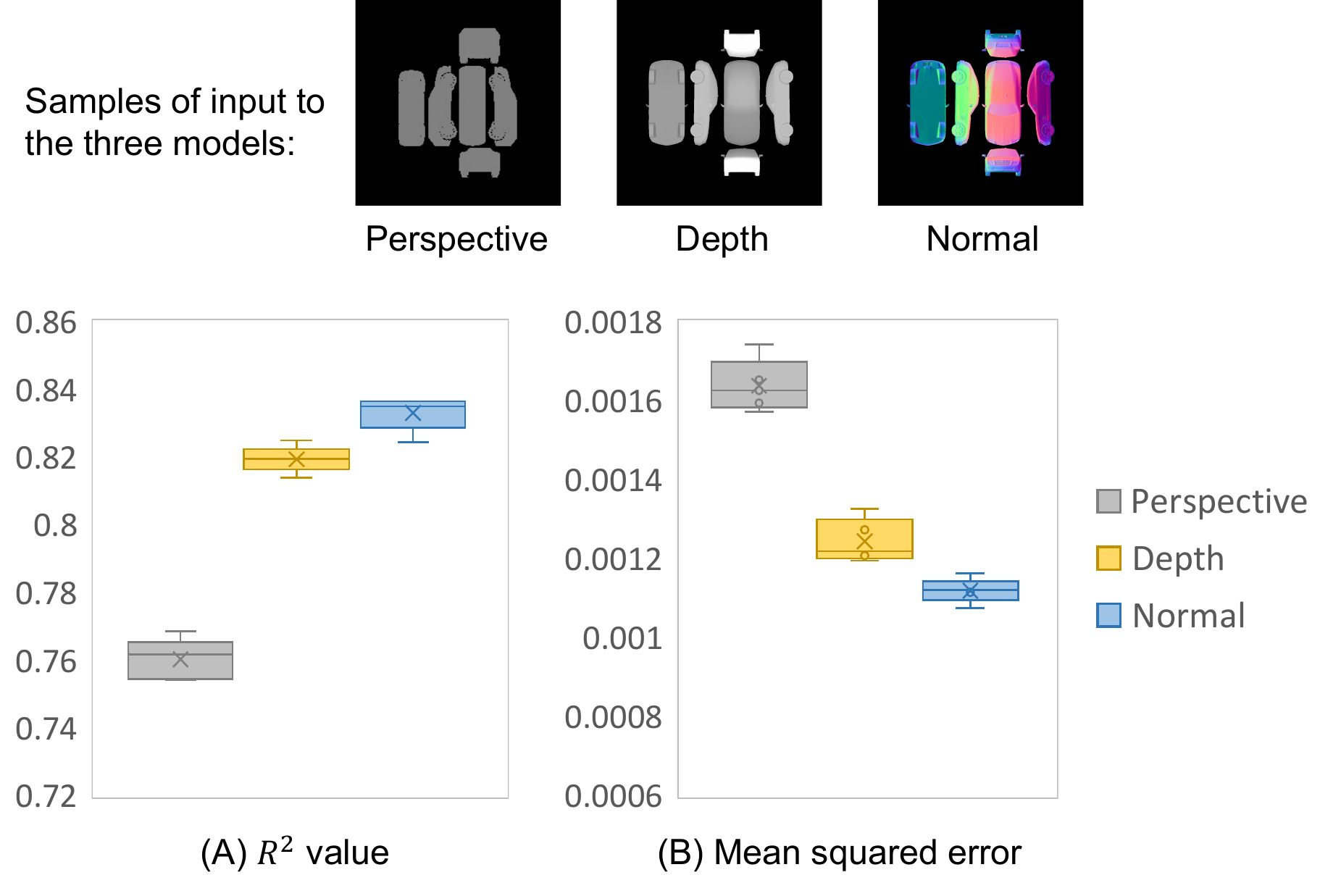}
    \caption{The performances of the surrogate models using the proposed representation and the commonly used perspective renderings, respectively}
    \label{fig:common}
\end{figure}

The proposed representation is also more informative than the perspective renderings as input for car drag coefficient evaluation. In this set of experiments, the attn-ResNeXt model takes the 2D perspective renderings and the proposed normal and depth renderings as input, respectively. Figure~\ref{fig:common} depicts their $R^2$ and MSE values. The models using the normal renderings and depth renderings achieve significantly higher $R^2$ values and lower MSE values than that using the 2D perspective renderings. That is, the normal and depth information conveyed by the proposed representation enables the model to capture more informative features for drag coefficient prediction. 

Additionally, the normal renderings are more informative than the depth renderings for this task when used separately. Two possible reasons can explain this. First, the normal renderings reflect the surface features directly, while the depth renderings provide the positional information from which the surface features can be inferred in a less straightforward way. Since the aerodynamic performance of a car is determined by its surface features, this difference probably makes the normal renderings more informative for drag coefficient prediction. Second, the three color channels of the normal renderings store different information regarding the normal vectors along the $x$, $y$, and $z$ coordinates, respectively. In comparison, the three color channels of the depth rendering store the same information regarding the distance between the camera and a certain point. The richness of the color channels may also allow the surrogate models to capture more information from the normal renderings.

Then, we compare our surrogate model with the baseline models from two prior studies. The first study~\cite{Gunpinar2019ADynamics} ran 2D CFD simulations with car silhouettes. The second study~\cite{Umetani2018LearningDesign} ran 3D CFD simulations with simplified car designs with certain detailed features (e.g., wheels and mirrors) removed. Moreover, each car in their dataset was only simulated for 10 seconds, which might not return converged and reliable simulation results. That is, their simulations are rough compared to ours. Both baseline models employed parametric car representations to predict the simulated drag coefficients. As shown in Table~\ref{tab:comparison}, this study has advantages over the two baseline studies from three perspectives. First, unlike the other two studies targeting at simplified car designs, this study aims to predict the drag coefficients of full-featured car designs. The authenticity of the cars in our dataset makes the associated surrogate model more applicable to the practical car design process. Second, since our dataset covers different types of cars from a wider range of drag coefficients, our surrogate model trained on it is more likely to be generalized to different car categories. Third, our model achieves a lower MSE compared to the first model and a comparable average prediction error with the second model. Since our dataset covers a wider drag coefficient range, the MSE and average error of our model could be lower when it is tested within their drag coefficient ranges, as shown in Figure~\ref{fig:prediction}.

\begin{table*}[t]
\caption[Table]{The comparison between the best model from this study and two prior studies: Study 1-\cite{Gunpinar2019ADynamics} and Study 2-\cite{Umetani2018LearningDesign}}\label{tab:comparison}
\centering{%
\begin{tabular}{lccc}
\toprule
 & \textbf{Ours} & \textbf{Study 1} & \textbf{Study 2}\\
\hline
Input to CFD Simulation & 3D meshes & 2D silhouettes & 3D meshes \\
Input to surrogate models & 2D normal and depth renderings & Parametric & Parametric \\
Authenticity & Original & Simplified & Simplified \\
Drag Coefficient Range & 0.17-0.85 & 0.21-0.51 & 0.2-0.6 \\
Mean Squared Error & $8.2\times10^{- \ 4}$ & $1.84\times10^{- \ 3}$ & \\
Average Error & 0.024 & & 0.013-0.021 \\
\bottomrule
\end{tabular}
}
\end{table*}

The above comparisons verify the effectiveness of the proposed representation. The proposed representation integrating six single-view renderings contains more comprehensive geometric information than any single-view rendering. Moreover, the proposed representation is more informative than the 2D perspective renderings for two reasons. First, the proposed normal and depth renderings convey the geometric information regarding the surface normal and positional features of each point of a 3D shape. Second, the orthographic projection used to generate the proposed representation avoids geometric distortion compared to the perspective projection. These advantages of the proposed representation allow us to reconstruct 3D shapes from them without any learning process, while it is challenging to accurately reconstruct 3D shapes from the 2D perspective renderings without a learning process. Moreover, the proposed representation method is generalizable to broader 3D shape categories whose major geometric information can be captured from the six orthographic views, such as airplanes, ships, bottles, chairs, and so forth.

The proposed 2D representation has the potential to promote 3D shape generation, evaluation, and optimization using deep learning models. The 3D representations of 3D shapes are either sparse or redundant in many cases. For example, only surface information is needed to represent a car body design. When it is represented as voxels, all the voxels inside the surface are redundant. The redundancy and sparsity of 3D representations make it highly computationally expensive to learn 3D shapes. With limited computational power, deep learning models struggle to handle high-resolution 3D shapes represented by voxels, meshes, or point clouds. Accordingly, these models do not allow for the generation, evaluation, and optimization of 3D shapes with plenty of geometric details, hindering their applications to real-world problems. Moreover, as AI technologies are more explored to handle 2D data as of now, the proposed 2D representation enables us to handle 3D shapes with more powerful 2D AI technologies. It is much easier and less expensive to increase the resolution of the 2D representation of 3D shapes than doing that with the 3D representations directly. Therefore, the proposed representation is promising to enable 3D shape generation, evaluation, and optimization at a higher resolution with less computational power needed.

\subsection{Limitations and Future Work}
While the proposed 2D representation, dataset, and surrogate model are promising, they have limitations and leave room for further improvement. First, the proposed 2D representation is insufficient to model more complex geometric structures, such as lattice cubes and flowers. Moreover, although the proposed representation is informative for machine learning, it is less intuitive for human perception compared to 3D representations, such as meshes and point clouds. Second, the dataset introduced in this paper is far smaller than the training sets typically used for deep learning models. In particular, the number of samples with high drag coefficients is low. The small dataset leads to significant over-fitting during the training process. Our hope is to expand this dataset with help from the community. We aim to improve and verify its reliability by training and testing the developed model with a sizable dataset. Third, while we show that the integrated renderings are more informative than the single-view renderings, alternative integration techniques may be more effective. We will explore such alternatives in future work. Fourth, the approach proposed in this paper for drag coefficient prediction is a purely data-driven approach, which does not leverage any physics knowledge regarding CFD simulations. The performance of the surrogate model depends on the quality and quantity of the data, and is unlikely to perform well on inputs far from the training set. We will attempt to incorporate physics into our surrogate model in future work. Lastly, the surrogate model developed in this paper can only make predictions using the proposed 2D representations of cars and does not apply to common car images. A promising future direction is to associate the proposed 2D representations with real images so that the surrogate model can make predictions using easily accessible car images.

\section*{CONCLUSION}
Drag coefficient evaluation is an indispensable element of the aerodynamic design of cars, which has a critical influence on car fuel efficiency. In this paper, we develop a surrogate model that enables accurate, fast, and differentiable drag coefficient evaluation. This surrogate model is built on a new two-dimensional (2D) representation of three-dimensional (3D) shapes. This representation embeds depth and surface normal information into 2D renderings and combines information from six orthographic views. The results of this study suggest that our proposed representation is more effective and informative than simple 2D perspective renderings for drag coefficient prediction. To train our model, we also assemble a diverse dataset of high-quality 3D car meshes labeled by their drag coefficients, as computed by computational fluid dynamics (CFD) simulations. This dataset, upon public release, can drive the development of other data-driven design approaches. In total, our contributions facilitate the data-driven design of 3D aerodynamic cars and can be readily combined with generative AI techniques to automate design creation.

%%%%% Acknowledgments %%%%%%%%%%%%%%%%%%%%%%%%%%%
\section*{Acknowledgments}
This research was supported in part by the Toyota Research Institute. Additionally, we thank Mr. Hanqi Su for helping us select high-quality car meshes from ShapeNet.

\bibliographystyle{unsrtnat}
\bibliography{references}  %%% Uncomment this line and comment out the ``thebibliography'' section below to use the external .bib file (using bibtex) .





% \end{thebibliography}

\end{document}